\documentclass[10pt,twocolumn,letterpaper]{article}

\usepackage{cvpr}
\usepackage{times}
\usepackage[bold]{hhtensor}
\usepackage{epsfig}
\usepackage{graphicx}
\usepackage{amsmath}
\usepackage{amssymb}
\usepackage{subfig}
\usepackage{authblk}

\newcommand{\squishlist}{
	\begin{list}{$\bullet$}
		{ \setlength{\itemsep}{0pt}
			\setlength{\parsep}{1pt}
			\setlength{\topsep}{1pt}
			\setlength{\partopsep}{0pt}
			\setlength{\leftmargin}{1.5em}
			\setlength{\labelwidth}{1em}
			\setlength{\labelsep}{0.5em} } }
	
	\newcommand{\squishend}{
	\end{list}  }

\usepackage[breaklinks=true,bookmarks=false]{hyperref}

\cvprfinalcopy

\begin{document}

\title{MIML-FCN+: Multi-instance Multi-label Learning via Fully Convolutional Networks with Privileged Information}

\author[*]{Hao Yang}
\author[**]{Joey Tianyi Zhou}
\author[*]{Jianfei Cai}
\author[*]{Yew Soon Ong}

\affil[*]{School of Computer Science and Engineering, NTU, Singapore. \url{Lancelot365@gmail.com}}


\maketitle

\begin{abstract}
	
	Multi-instance multi-label (MIML) learning has many interesting
	applications in computer visions, including multi-object
	recognition and automatic image tagging. In these applications,
	additional information such as bounding-boxes, image captions and
	descriptions is often available during training phrase, which is
	referred as privileged information (PI). However, as existing
	works on learning using PI only consider instance-level PI
	(privileged instances), they fail to make use of bag-level PI
	(privileged bags) available in MIML learning. Therefore, in this
	paper, we propose a two-stream fully convolutional network, named
	\textsc{MIML-FCN+}, unified by a novel PI loss to solve the
	problem of MIML learning with privileged bags. Compared to the
	previous works on PI, the proposed \textsc{MIML-FCN+} utilizes the
	readily available privileged bags, instead of hard-to-obtain
	privileged instances, making the system more general and practical
	in real world applications. As the proposed PI loss is convex and
	SGD-compatible and the framework itself is a fully convolutional
	network, \textsc{MIML-FCN+} can be easily integrated with
	state-of-the-art deep learning networks. Moreover, the flexibility
	of convolutional layers allows us to exploit structured
	correlations among instances to facilitate more effective training
	and testing. Experimental results on three benchmark datasets
	demonstrate the effectiveness of the proposed \textsc{MIML-FCN+},
	outperforming state-of-the-art methods in the application of
	multi-object recognition.
\end{abstract}

\section{Introduction}

In the traditional supervised learning, each training instance is
typically associated with one label. With the rapid development of
deep learning~\cite{Alex2012}, such single-instance single-label
classification problem is nearly solved, given abundant
well-labelled training data. For example, for single object
recognition tasks, such as ILSVRC, several methods have already
achieved super-human performance~\cite{He2015,He2016,Ioffe2015}.
However, in many real-world applications, instead of training
instances, we often encounter the problem of training bags, each
of which usually contains many instances, e.g., frames in a video
clip, object proposals of an image, which is referred as
multi-instance setting. In addition, to accurately describe a bag,
we often need to associate multiple labels or tags to it, which is
referred as multi-label setting. Such multi-instance multi-label
(MIML) learning setting~\cite{Zhou2012} is more general, but more
challenging.

\begin{figure}
	\centering
	{\includegraphics[width=0.45\textwidth]{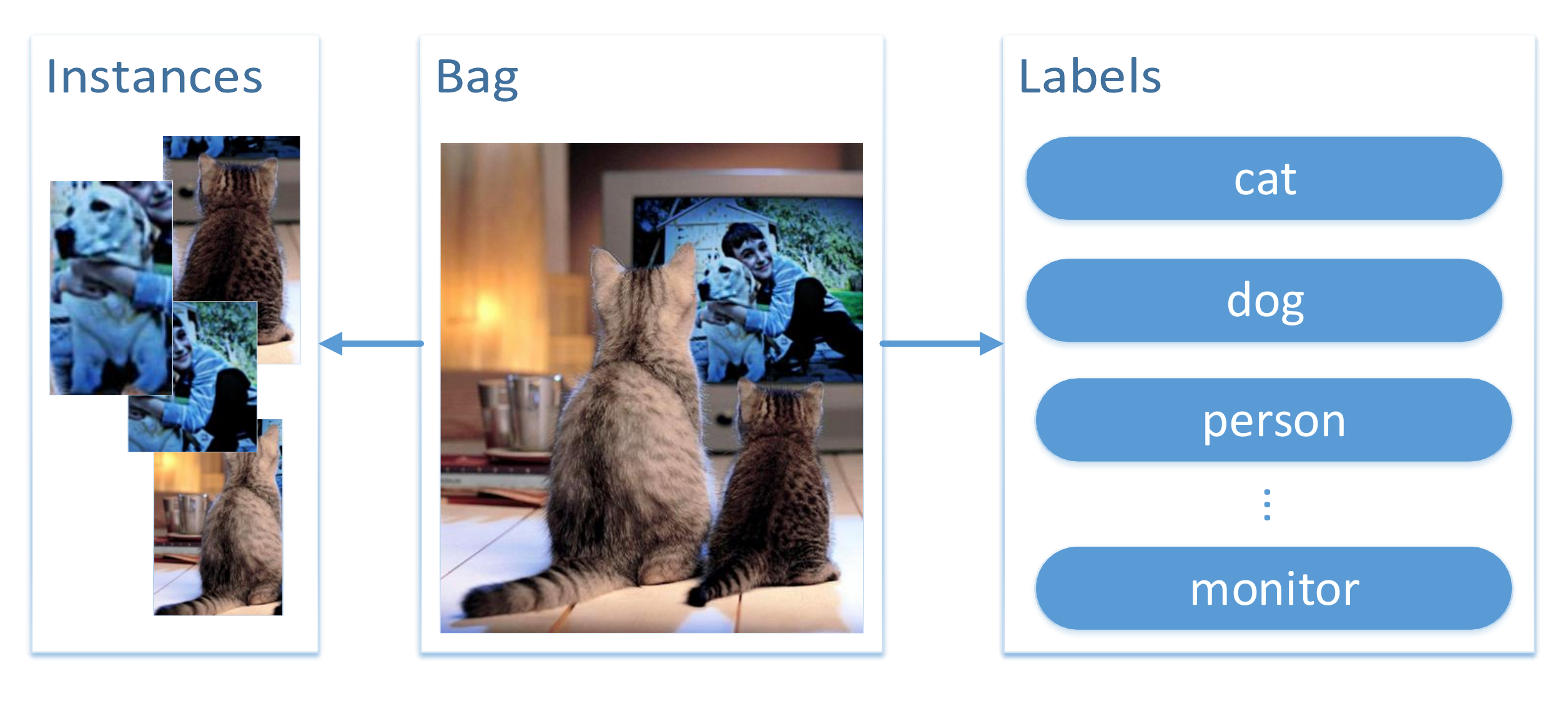}}
	\caption{A practical example of multi-instance multi-label (MIML) learning problem. Here we consider the image as a bag, the proposals extracted from the image as instances, and the objects contained in the image as bag labels.}
	\label{example-miml}
\end{figure}

MIML learning has many applications in computer vision. For
example, in multi-object recognition and automatic image tagging
problems, an image can be decomposed into many object proposals,
where we can treat each image as a bag and each of its proposals
as an instance in the bag, as illustrated in
Fig.\ref{example-miml}. The MIML learning problem essentially is,
given training bags with only bag-level labels, how to learn an
effective model that can accurately assign multiple labels to new
bags. MIML learning problems have attracted significant attentions
in the past few
years~\cite{Simonyan2014,Yang2016a,Oquab2014,Bilen2016}. With the
release of large scale multi-label datasets such as
YFCC100M~\cite{YFCC100M} and Google Open Images~\cite{openimages},
it will stimulate more large-scale MIML learning studies.

On the other hand, in many applications, additional information is
often available in the \emph{training} phrase. Vapnik and
Vashist~\cite{Vapnik2009} referred such additional information as
privileged information (PI) and showed that PI can be utilized as
a teacher to train more effective models in traditional supervised
learning problems. This motivates us to incorporate PI into MIML
learning problem. However, there are two main obstacles hinging us
from applying learning using privileged information (LUPI)
paradigm to MIML problems.

First of all, existing works on privileged information only
consider instance-level
PI~\cite{Vapnik2009,Vapnik2015a,Vapnik2015b,Li2014}. This might
not be a problem for traditional supervised learning, but for most
MIML tasks, instance-level PI, where each training instance in
each training bag must have a corresponding privileged instance,
is hard to obtain. In contrast, bag-level PI is much easier to
acquire, and is often already available. Take the aforementioned
multi-object recognition problem as an example. It is hard to
obtain privileged information for each object proposal we extract,
but for images, there are bounding-boxes, captions, descriptions,
which can all be used as bag-level PI. Another example could be
video recognition, where each clip can be viewed as a bag and
frames or sub-clips in each clip, containing different objects,
activities, can be viewed as instances in the bag. It is clear
that bag-level PI such as video descriptions are much easier to
obtain. Therefore, in MIML learning with privileged information,
it is more general and meaningful to consider bag-level PI, which
is lacking in the current literature.

Secondly, most existing works on PI are still based on the
original SVM+ formulation, where PI is used as slack functions.
Although this formulation has many theoretical and practical
merits~\cite{Vapnik2015a}, it is hard to incorporate it into
state-of-the-art deep learning paradigm in an end-to-end fashion
as the SVM+ formulation is not stochastic gradient descent (SGD)
compatible. Thus, existing PI works fail to benefit from rapid
developments of deep learning.

In this paper, we address these two problems by proposing a
two-stream fully convolutional network, which we refer as
\textsc{MIML-FCN+}. In the proposed framework, each stream handles
one source of information, namely training bags and privileged
bags, respectively. The two-stream networks are unified by a novel
PI loss, which follows the high level idea of
SVM+~\cite{Vapnik2009} but with a totally different realization
oriented for deep learning. Specifically, we propose to utilize
privileged bags to model training losses and use it as a convex
regularization term, which facilitates SGD-compatible loss and
end-to-end training. In addition, motivated by the
work~\cite{Zhou2009}, which shows exploiting structured
correlations among instances can help MIML learning, we further
propose to construct a graph for each bag and incorporate the
structured correlations into our MIML-FCN+ framework, thanks to
the structure of fully convolutional networks, where filter sizes
and step sizes of the convolutional layers can be easily adjusted.

The major contributions of this paper are threefold. First, we
propose and formulate a new problem of MIML learning with
privileged bags, which is a much more practical setting in real
world applications. To the best of our knowledge, this is the
first work exploiting privileged bags instead of privileged
instances. Second, we propose a two-steam fully convolution
network with a novel PI loss, MIML-FCN+, to solve the MIML+PI
learning problem. Our solution is fully SGD-compatible and can be
easily integrated with other state-of-the-art deep learning
networks such as CNN and RNN. Our MIML-FCN+ is flexible to combine
different types of information, e.g. images as training bags and
texts as privileged bags. It can also be easily extended to make
use of privileged instances if available. Third, we further
propose a way to incorporate graph-based inter-instance
correlations into our MIML-FCN+.

\section{Related Works}
\textbf{Multi-instance Multi-label Learning}: During the past
decade, many MIML algorithms have been
proposed~\cite{Luo2010,Zhou2006,Zhou2012,Nguyen2010,Nguyen2013}.
For example, MIMLSVM~\cite{Zhou2006} degenerates the MIML problem
into solving the single-instance multi-label problem while
MIMLBoost~\cite{Zhou2006} degenerates MIML into multi-instance
single-label learning, which suggest that MIML is closely related
to both multi-instance learning and multi-label learning. Ranking
loss had been shown to be effective in multi-label learning, and
thus Briggs et al.~\cite{Briggs2012} proposed to optimize ranking
loss for MIML instance annotation. In terms of generative methods,
Yang et al.~\cite{Yang2009} proposed a Dirichlet-Bernoulli
alignment based model for MIML learning problem. In contrast, in
this work we consider using privileged information to help MIML
learning under the deep learning paradigam, which has not been
explored before.

Many computer vision applications such as scene classification,
multi-object recognition, image tagging, and action recognition,
can be formulated as MIML problelms. For instance, Zha et
al.~\cite{Zha2008} proposed a hidden conditional random field
model for MIML image annotation. Zhou et al.~\cite{Zhou2006}
applied MIML learning for scene classification. Several
works~\cite{Oquab2014,Yang2016a,Bilen2016} also implicitly
exploited the MIML nature of multi-object recognition problem.

\textbf{Learning Using Privileged Information (LUPI):} LUPI
assumes there are additional data available during training, i.e.
privileged information (PI), which are not available in testing.
Vapnik and Vashist~\cite{Vapnik2009} proposed an SVM+ formulation
that exploits PI as slack variables during training to ``teach"
students to learn better classification model. The idea was later
developed into two schemes: similarity control and knowledge
transfer~\cite{Vapnik2015a}. LUPI has also been utilized in metric
learning~\cite{Fouad2013}, learning to rank~\cite{Sharmanska2013}
and multi-instance learning~\cite{Li2014}. A few works have
applied PI to computer vision applications. For example, Li et
al.~\cite{Li2014} applied PI for web images recognition.
Sharmanska et al.~\cite{Sharmanska2013} applied PI for image
ranking and retrieval. However, most of the existing PI works
consider only instance-level PI, are still based on SVM+
formulation, which is hard to be incorporated into a deep learning
framework in an end-to-end fashion. In this work, we address all
these limitations by a two-stream fully convolutional network and
a new PI loss.

\section{Proposed Approach}
In the context of multi-instance and multi-label (MIML) learning,
assume there are $n$ bags in the training data, denoted by
$\{X_i,Y_i\}_{i=1}^{n}$, where each bag $X_i$ has $m_i$ instances
$\{\vec{x}_{i,j}\}_{j=1}^{m_i}$ and $Y_i$ contains the labels
associated with $X_i$. We represent $Y_i$ as a binary vector of
length $C$, where $C$ is the number of labels. The $k$-th
dimension $Y_i(k) = 1$ if $k$-th label $c_k$ is associated with at
least one instance in $X_i$; otherwise $Y_i(k) = -1$. In other
words, denoting $\vec{y}_{i,j}$ as the label vector of instance
$\vec{x}_{i,j}$, $Y_i(k) = 1$ if and only if $\exists j,
\vec{y}_{i,j}(k) = 1$. Note that in common MIML setting, the
instance-level labels $\vec{y}_{i,j}$ are usually assumed not
available.

In learning using privileged information (LUPI) paradigm, we
further assume that for each training bag, there exists a
privileged bag $X_i^*$. $X_i$ and $X_i^*$ are two views of the
same real world image.  $X_i^*$ can contain $m_i^*$ instances
$\{\vec{x}_{i,j}^*\}_{j=1}^{m_i^*}$. Here $m_i^*$ is generally
different with $m_i$, and there is no instance-level
correspondence between training data and privileged information.
This is one fundamental difference between our work and previous
LUPI studies that always assume each training instance
$\vec{x}_{i,j}$ has a corresponding privileged instance
$\vec{x}_{i,j}^*$.

\subsection{MIML Learning through FCN}
\label{FCN}

\textbf{MIML:} We start with reviewing the general MIML learning
pipeline. Given a bag $X$, the goal of MIML learning is
essentially to learn a model $F(X)$ such that the difference
between $F(X)$ and the true label $Y$ is small. An MIML system
$F(\cdot)$ generally consists of two components: a non-linear
feature mapping component and a classification component. In the
feature mapping component, each $d$-dimensional training instance
$\vec{x}$ is mapped from the input space to the feature space,
where training data could be linearly separable, by a non-linear
mapping function $\phi(\cdot)$.

In the classification component, each instance is first mapped
from the feature space to the label space by
\begin{equation}
f(\vec{x}) =  \phi(\vec{x}) W,
\end{equation}
where $W$ is a $d'\times C$ weight matrix classifying the $d'$-dim
mapped instance $\phi(\vec{x})$ to a label vector. Then, the
predicted instance-level labels is transferred to the bag-level
labels. According to the MIML learning definition, the relation
between instance-level labels $\vec{y}_{j}$ and bag-level labels
$Y$ can be expressed either as:
\begin{equation}
\label{max-relation} Y = \max_j(\vec{y}_{j}),
\end{equation}
where $\max$ is the per-dimension max operation, or alternatively
as a set of linear constraints~\cite{Andrews2002}:
\begin{equation}
\left\{\begin{matrix}

\sum_{j} {\frac{\vec{y}_{j}(k) + 1}{2}} \geq 1 && \text{ if } Y(k) = 1,\\
\vec{y}_{j}(k) = -1, \forall j&& \text{ if } Y(k) = -1.

\end{matrix}\right.
\end{equation}
Let us consider the first case, i.e., using
Eq.(\ref{max-relation}) to map instance-level labels to bag-level
labels. With this relation, the bag-level label prediction becomes
\begin{equation} \label{eq:FX}
F(X) = \max_{\vec{x}\in X} \phi(\vec{x}) W .
\end{equation}
Thus, the objective function for MIML learning can be written as
\begin{equation}
\label{obj1}
\begin{matrix}
\min & \mathcal{L}(Y, F(X)),
\end{matrix}
\end{equation}
where $\mathcal{L}(\cdot)$ is a suitable multi-label loss such as
square loss or ranking loss.

\textbf{MIML-FCN:} It is not difficult to see that the above
formulated MIML learning can be realized via a neural networks.
First, in terms of feature mapping, the previous MIML studies
usually project the data from input space into feature space by
pre-defined project functions, such as kernels~\cite{Andrews2002}
and Fisher vectors~\cite{Wei2014}, or learned linear
projections~\cite{Huang2014}, which are incompatible with neural
networks. On the other hand, the combination of multiple layers of
fully connected layers and non-linear activation functions has
proven to be a powerful non-linear feature mapping~\cite{Cho2009,
	Perronnin2015}. Thus, in our framework, we employ multiple
convolutional layers and ReLU layers as our feature mapping
component. The reason that we use fully convolutional networks
(FCN) without including any fully connected layers is that FCN is
more flexible and can handle any spatial resolution~\cite{Oquab2014},
which is needed for the considered MIML problem since the number
of instances in each bag varies.

Particularly, with $\phi_l(\vec{x}) = g(\vec{x}, W_l) + \vec{b}_l$
denoting the $l$-th convolutional layer, where $\vec{x}$ is the
input, $g$ is the convolution operation, $W_l$ is the parameters
and $\vec{b}_l$ is the bias, and $\sigma(\cdot)$ denoting the
non-linearity, the feature mapping component $\phi$ of our
framework can be expressed as:
\begin{equation}
\phi(\vec{x}) =  \sigma(\phi_L(\dots
\sigma(\phi_2(\sigma(\phi_1(\vec{x}))))\dots)),
\end{equation}
if there are in total $L$ layers. For $1\times 1$ filters, the
convolution operator $g$ is just a dot-product.

\begin{figure}
	\centering
	{\includegraphics[width=0.45\textwidth]{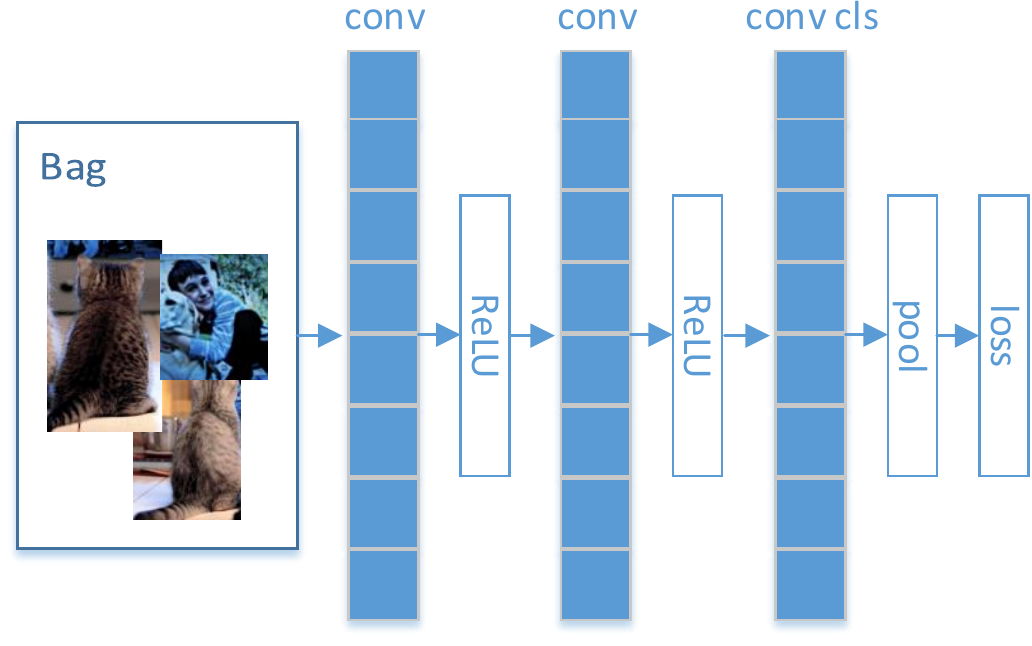}}
	\caption{An example of our proposed MIML-FCN architecture. The input is a bag of $m_i$ instances, typically organized as $1\times m_i\times d$ (The feature dimension / channel $d$ is usually omitted for simplicity). MIML-FCN typically contains a few layer-pairs (e.g. 2 layer-pairs here) of $1\times 1$ conv layer and ReLu layer for feature mapping, one $1\times 1$ conv layer for classification, one global pooling layer (e.g. max pooling here) and one loss layer.}
	\label{example-fcn}
\end{figure}

Other operations in MIML can also be easily mapped into FCN.
Specifically, the classification component in~\eqref{eq:FX} is
realized by a convolutional layer with $1 \times 1$ filter size
and parameters $W$ to project the learned feature to the label
space,  followed by a pooling layer to extract per-bag prediction.
The loss function in~\eqref{obj1} is realized by a loss layer with
appropriate SGD-compatible multi-label loss such as square
loss~\cite{Wei2014,Yang2016b} and ranking
loss~\cite{Wei2014,Yang2016a}. Fig.~\ref{example-fcn} shows an
example of our proposed MIML-FCN architecture, which typically
consists of a few layer-pairs (e.g. 2 layer-pairs here) of
$1\times 1$ conv layer and ReLu layer for feature mapping, one
$1\times 1$ conv layer for classification, one global pooling
layer (e.g. max pooling here) and one loss layer.

We would like to point out that similar network structure has been
used in several previous works on multi-object recognition and
weakly supervised object detection~\cite{Oquab2014,Bilen2016},
while we explicitly use such structure for MIML and more
importantly we will extend it to incorporate privileged
information as well as structured correlations among instances.


\subsection{MIML-FCN with Privileged Bags}
\label{FCN+} Training the proposed MIML-FCN might not be as easy
and straightforward as training a single-label CNN, as the MIML
learning itself is by definition non-convex. As a result, the
framework might not reach optimal classification accuracies even
if the hyperparameters are carefully tuned. Fortunately, in many
applications there often exists additional information, referred
as privileged information (such as image captions in multi-object
recognition), in training stage that can help us learn a better
model.

\textbf{SVM+:} Learning using privileged information (LUPI)
paradigm was first introduced by Vapnik and
Vashist~\cite{Vapnik2009}. They utilized privilege information as
the slack variables in the SVM formulation, called SVM+.
Specifically, their (linear) SVM+ objective function is:
\begin{equation}
\label{svm+}
\begin{matrix}
\underset{\vec{w},b,\vec{w}^*,b^*}{\min} & \frac{1}{2}(\|\vec{w}\|+\gamma\|\vec{w}^*\|) + C\sum_{j=1}^{n}\xi(\vec{x}_j^*) \\
\text{s.t. } & y_j(\vec{w}\vec{x_j}+b) \geq 1 - \xi(\vec{x}_j^*),
\text{ } \xi(\vec{x}_j^*) \geq 0, \text{ } \forall i,
\end{matrix}
\end{equation}
where $\gamma$ and $C$ are the trade-off parameters,
$\vec{w}\vec{x_j}+b$ is the classification model,
$\xi(\vec{x}_j^*) = \vec{w}^{*}\vec{x_j}^*+b^*$ is the slack
function, replacing the slack variables $\xi_j$ in the original
SVM formulation. This slack function acts as a teacher by
correcting the concepts of similarity of original training data by
privileged information during training process.

Although LUPI paradigm has many good theoretical and practical
merits~\cite{Vapnik2009, Vapnik2015a, Vapnik2015b}, directly
applying this formula to MIML learning setting is not plausible
due to two main problems. Firstly, in most MIML problems,
instance-level PI, or privileged instances, is difficult to
obtain. The previous work~\cite{Li2014} that extends SVM+ directly
to MISVM+ requires privileged instances, which greatly limits its
applicable areas. In contrast, bag-level privileged information,
or privileged bags, is much easier to get and often readily
available. Secondly, Eq.(\ref{svm+}) is relatively difficult to
solve compared to traditional SVM. Although there are efforts on
developing new dual coordinate descent algorithm to improve the
training efficiency~\cite{Li2016c}, unifying LUPI and deep
learning in an end-to-end fashion is still not tackled.

\textbf{MIML-FCN+:} To overcome the obstacles, we construct a
two-stream network, named \textsc{MIML-FCN+}. The first stream
models training bags (same as MIML-FCN), and the second stream
models the privileged bags. With this configuration, our framework
not only effectively utilizes privileged bags, but also allows the
flexibility to deal with different types of data. For instance, if
the training bags are images and privileged bags are texts, we
clearly need to map these data to different feature spaces in
order to effectively extract knowledge, for which our two-stream
networks can be configured accordingly. We could even employ RNN
if the privileged information is text.

With \textsc{MIML-FCN+}, we need an SGD-compatible PI loss to
replace the original loss so that we can utilize privileged bags
as ``teachers" during training. Since dealing with slack variables
is difficult, inspired by the high level idea
of~\cite{Vapnik2009}, we propose to utilize privileged information
to model the loss of training data, penalize the difference of PI
modelled loss and true loss, and add the difference as a
regularization term to Eq.(\ref{obj1}).

Specifically, assume that for each training bag $X_i$, we have a
privileged bag $X_i^*$. We use a second stream of network (called
slack-FCN) to model privileged bags. Compared to the first stream
of network (called loss-FCN), which models the training bags, the
goal of the second stream is not to learn a classification model,
but to model the loss of the first stream. Denote the output of
the second stream for an input privileged bag $X^*$ as $F^*(X^*)$,
the two streams share the same loss layer defined by:
\begin{equation}
\label{2fcn-obj}
\begin{matrix}
\min & \mathcal{L}(Y, F(X)) + \lambda\|\mathcal{L}(Y, F(X)) -
F^*(X^*)\|_2^2,
\end{matrix}
\end{equation}
where $\|\cdot\|_2$ is the L$2$ norm.

In SVM+, privileged information is used to model slack variables,
which can be viewed as a set of tolerance functions that allows
the margin constraints to be violated. In the proposed
\textsc{MIML-FCN+}, we make use of this idea and utilize
privileged information to approximate classification error of
original training data. On one hand, slack-FCN models the
difficulty of classifying training bags with privileged
information. On the other hand, slack-FCN can provide a way to
regularize the classification errors to avoid over-fitting.

The proposed \textsc{MIML-FCN+} can be optimized in an alternating
fashion. Specifically, we update the loss-FCN while fixing the
parameters of slack-FCN until it converges, and subsequently
update slack-FCN while fixing the parameters of loss-FCN. This
process is repeated for several times until the whole system
converges.

\subsection{Utilizing Structured Correlations among Instances}
\label{graph} In the previous sections, we treat instances in a
bag as independently and identically distributed (i.i.d) samples
by using $1 \times 1$ filter in the convolutional layers. The
assumption ignores the fact that instances in a bag are rarely
independent , and correlations among instances often contain
structured information. Considering object proposals from an image
as an example, these proposals are clearly correlated as there
exist large overlaps among them. Zhou et al.~\cite{Zhou2009}
showed that treating instances as non-i.i.d samples could be
helpful for learning more effective classifier. Their MIGraph and
miGraph methods explicitly or implicitly use graph to exploit the
structured correlations among instances in each bag.

Our MIML-FCN+ framework is flexible to incorporate such structured
correlations among instances since our framework is based on FCN,
where the filter sizes of convolutional layers can be easily
adjusted to accommodate graph input. Specifically, we first
construct a Nearest-Neighbour~(NN) graph for each bag, which is a
simple and effective way to capture correlations among instances
in each bag. Assume for each vertex in the graph, i.e., each
instance, there exist $k$ edges connecting to other vertices,
i.e., its $k$ nearest neighbours. We can organize this graph as a
$3$D tensor and use it as the input to our system. The
dimensionality of the tensor will be $k \times m_i \times d$,
where $m_i$ is the number of instances in bag $X_i$, and $d$ is
the dimension of each instance. Instead of using $1\times 1$filter
for the first convolutional layer, we use $k\times 1$ filters. In
this way, we essentially utilize not only each instance itself,
but also its $k$ nearest neighbours in the graph. By treating each
instance as a connected vertex in the graph, we could potentially
learn a more robust network.

\section{Multi-object Recognition: A Practical Example}
\label{example} In this section, we use multi-object recognition
as a practical example to show how to apply our proposed MIML-FCN+
framework. We also validate the performance of the proposed
\textsc{MIML-FCN+} on this application in the experiment section.

Multi-object recognition refers to recognizing multiple objects
from one single image. As the objects could be from different
locations, scales and categories, it is natural to extract object
proposals from training images. Thus, for training data, we refer
each image as a bag $X$ and feature extracted from the proposals
in the image as instances in the bag. Particularly, we utilize
ROI-pooled CNN features as features for proposals as
in~\cite{Ren2015}. We stack our MIML-FCN+ framework on top of
ROI-pooled CNN and train the entire system end-to-end.

\begin{figure}
	\centering
	\subfloat[Bounding boxes as
	PI]{\includegraphics[width=0.45\textwidth]{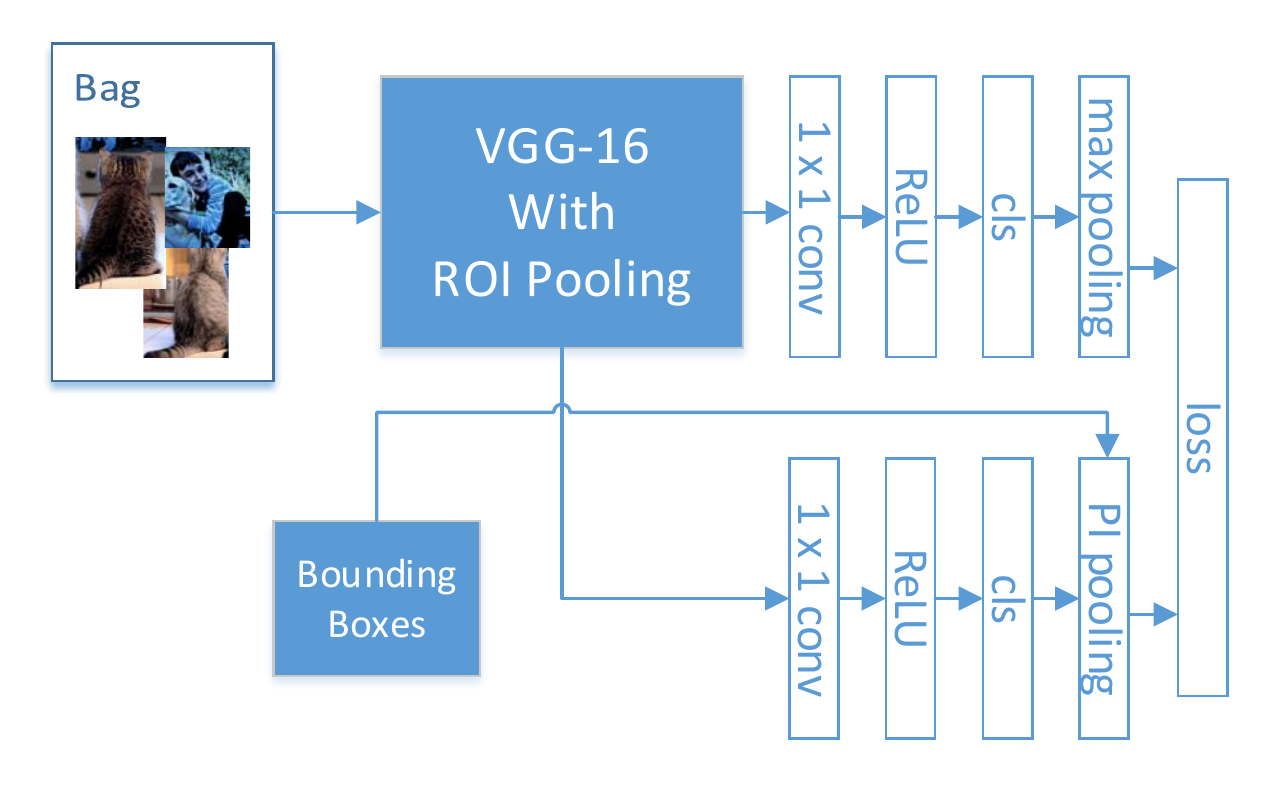}}\\
	\subfloat[Image Captions as PI]{\includegraphics[width=0.45\textwidth]{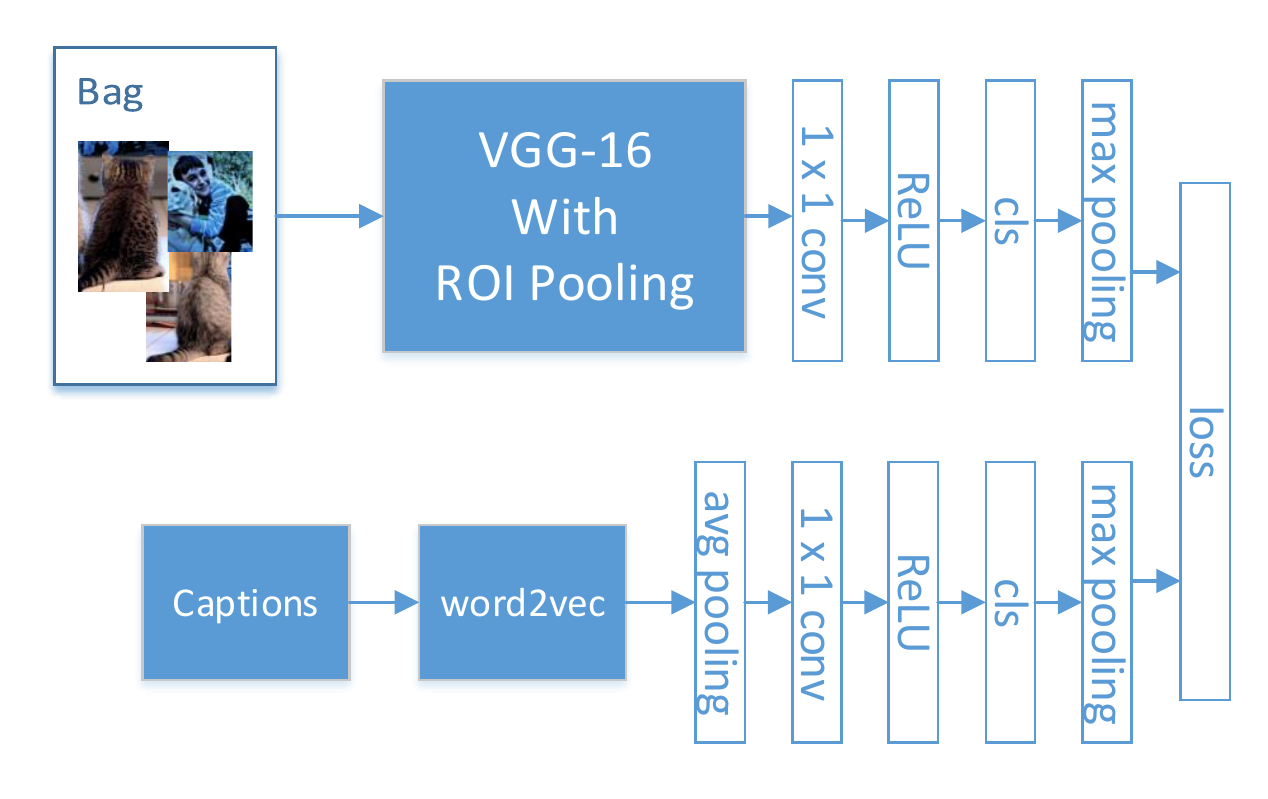}}
	\caption{The proposed \textsc{MIML-FCN+} combined with VGG-16 with ROI pooling for multi-object recognition. (a) with bounding boxes as PI; (b) with image captions as PI.}
	\label{pi}
\end{figure}

\textbf{Bounding boxes as PI:} For privileged bags, we utilize two
different types of privileged information. The first type of
privileged information is bounding boxes for objects. In order to
make of use of this information, we propose a PI pooling layer to
replace the global max pooling in the slack-FCN, as shown in
Fig.~\ref{pi}(a). This PI pooling layer identifies true positive
proposals that have $\geq 0.5$ IoU with ground truth bounding
boxes and average-pool the scores of these proposals so as to
better exploit the key instances in the bag. For negative
proposals, PI pooling layer sticks with max-pooling.
Mathematically, this PI pooling layer can be defined as:
\begin{equation}
F^*(k) = \left\{\begin{matrix}

\frac{1}{|P_k|}\sum_{j\in P_k} {\vec{\tilde{y}}^*_{j}(k)} && \text{ if } Y(k) = 1,\\
\max_j\vec{\tilde{y}}^*_{j}(k) && \text{ if } Y(k) = -1,
\end{matrix}\right.
\end{equation}
where $P_k$ is the set of proposals that have $\geq 0.5$ IoU with
ground truth bounding boxes of $k$-th category,
$\vec{\tilde{y}}^*_{j}(k)$ is the predicted instance or proposal
level scores in the slack-FCN for the $j$-th proposal and $k$-th
category, $F^*(k)$ is the predicted bag level score in the
slack-FCN for the $k$-th category, and $Y(k)$ is the corresponding
ground-truth for the loss-FCN.

Note that the proposed PI pooling can only be used in slack-FCN,
since it is only available in training but not in testing.
Considering only the pooling layer is changed in slack-FCN, both
loss-FCN and slack-FCN can share the same feature extraction
network, i.e. VGG-16 with ROI pooling as shown in
Fig.~\ref{pi}(a). Also, only one conv and Relu layer-pair is used
in both loss-FCN and slack-FCN for feature mapping, compared with
the two layer-pairs used in Fig.~\ref{example-fcn}. This is
because empirically we find one conv and Relu layer-pair performs
better.

\textbf{Image captions as PI:} The second type of privileged
information is image captions. Considering one image contains
multiple captions, we refer all captions of an image as a
privileged bag and each individual caption as one instance. To
better represent these captions, we extract work2vec features from
each word and use the weighted-averaged feature as the
representation for each sentence. Subsequently, we feed these
features into our slack-FCN, as shown in Fig.~\ref{pi}(b). Note
that it is also possible to use a RNN to encode each caption and
then append our slack-FCN, which will allow the whole system
end-to-end trainable.

We also need to decide what type of loss is suitable for training
the proposed networks for multi-object recognition. In this
research, we consider two losses: square loss and label ranking
loss.

\textbf{Square Loss:} The previous works~\cite{Wei2014,Yang2016a}
have shown that square loss can be a very strong baseline for
multi-label learning. Thus, we employ square loss as one
configuration for our framework. Specifically, the general cost
function in~\eqref{2fcn-obj} now becomes
\begin{equation}
\label{square}
\begin{matrix}
\min & \|Y -F(X)\|_2^2 + \lambda\|\|Y -F(X)\|_2^2 -
F^*(X^*)\|_2^2,
\end{matrix}
\end{equation}
for which the gradients with respect to $F(X)$ and $F^*(X^*)$ are
straightforward to compute.

\textbf{Label Ranking Loss:} Huang et al.~\cite{Huang2014}
proposed an approximated label ranking loss for the triplet
$(X,y,\bar{y})$, where $X$ is an input bag, $y$ is one of its
relevant labels, and $\bar{y}$ is one of its irrelevant label. The
key idea of this loss is to learn a model so that for every
training bag, its relevant labels rank higher than its irrelevant
labels by a margin. Specifically, the loss is defined
by~\cite{Huang2014}:
\begin{multline}
\label{label-rank}
\mathcal{L}_{r}(X,y,\bar{y}) = \epsilon(X,y)\left[1+F_{\bar{y}}(X)-F_y(X) \right]_+ \\
\approx\begin{cases}
0 & \text{ if { }}  \bar{y} \ \text{is not violated;}\\
S_{\bar{Y},v}(1+F_{\bar{y}}(X)-F_y(X)) &\text{otherwise}
\end{cases}
\end{multline}
where $S_{\bar{Y},v}$ is a normalization term~\cite{Huang2014}. To
train Eq.(\ref{label-rank}) in SGD-style, a triplet of
$(X,y,\bar{y})$ can be randomly sampled at  each iteration, and
the gradients of Eq.(\ref{label-rank}) can be easily calculated
and backpropagated.

For our \textsc{MIML-FCN+}, instead of the triplet
$(X,y,\bar{y})$, we sample a quadruplet  $(X,X^*,y,\bar{y})$ at
each iteration, and optimize:
\begin{equation}
\label{label-rank-obj}
\begin{matrix}
\min & \mathcal{L}_{r}(X,y,\bar{y}) +
\lambda\|\mathcal{L}_{r}(X,y,\bar{y}) -F^*(X^*,y,\bar{y})\|_2^2.
\end{matrix}
\end{equation}

Lastly, after training the proposed MIML-FCN+, we use only the loss-FCN during testing.

\section{Experiments}
\label{exp} In this section, we validate the effectiveness of the
proposed \textsc{MIML-FCN+} framework on three widely used
multi-label benchmark datasets.

\subsection{Datasets and Baselines}
We evaluate our method on the PASCAL Visual Object Calssess
Challenges (VOC) 2007 and 2012 datasets~\cite{VOC} and Microsoft
Common Objects in COtext (COCO) dataset~\cite{COCO}. The details
of these datasets are listed in Table~\ref{datasets}. We use the
train and validation sets of \textsc{VOC} datasets for training,
and test set for testing. For \textsc{MS COCO}, we use the
train2014 set for training, and val2014 for testing. For
\textsc{VOC} datasets, we use bounding boxes as privileged
information with the PI pooling layer as discussed
in~\ref{example}. For \textsc{MS COCO} dataset, we use two types
of PI, bounding boxes and image captions. The evaluation metric
used is average precision (AP) and mean average precision (mAP).

\begin{table*}
	\centering
	\caption{Dataset Information} \label{datasets}
	\begin{tabular}{c | r r r r r }
		\hline
		Dataset &\#Train Bags &\#Test Bags &\#Train Instances &\#Labels &\#Avg Labels \\ \hline
		\textsc{VOC 2007} &5011 &4952 &2.5M &20 &1.4\\
		\textsc{VOC 2012} &11540 &10991 &5.7M &20 &1.4\\
		\textsc{MS COCO} &82783 &40504 &41M &80 &3.5\\
	\end{tabular}
\end{table*}

We compare against several state-of-the-art methods for MIML
learning, \squishlist \item \textsc{MIMLfast}~\cite{Huang2014}: A
fast and effective MIML learning method based on approximate label
ranking loss as described in the previous section. MIMLfast first
projects each instance to a shared feature space with linear
projection, then learns $K$ sub-concepts for each label and
selects the sub-concept with maximum score. MIMLfast also employs
global max to obtain bag-level score. The main difference between
their method and our baseline MIML-FCN is that our feature mapping
can be non-linear. \item \textsc{miFV}~\cite{Wu2014}: A Fisher
vector (FV) based MIL learning method that encodes each bag to a
single Fisher vector, and then uses the ranking loss or square
loss to train a multi-label classifier on the FVs. \item
\textsc{RankLossSIM}~\cite{Briggs2012}: an MIML learning extension
of the ranking SVM formulation. \squishend

There exist other MIML learning methods such as MIMLSVM,
MIMLBoost~\cite{Zhou2006} and KISAR~\cite{Li2012}, but they are
too slow for our large-scale applications. Other than the MIML
learning methods, we also compare our \textsc{MIML-FCN+} framework
with the state-of-the-art approaches for multi-object recognition
that do not formulate the task as MIML learning problem, including
VeryDeep~\cite{Simonyan2014}, WSDDN~\cite{Bilen2016}, and the MVMI
framework~\cite{Yang2016a}. However, we did not compare with the
existing PI methods such as SVM+~\cite{Vapnik2009} and
sMIL+~\cite{Li2014}, since they can only deal with privileged
instances but not privileged bags. As far as we know, our proposed
\textsc{MIML-FCN+} is the only method that can make use of
privileged bags.

For our own \textsc{MIML-FCN+} framework, we consider three
different variations: \squishlist \item \textsc{MIML-FCN}: Basic
network without PI. \item \textsc{MIML-FCN+}: Two stream networks,
loss-FCN and slack-FCN, using either bounding boxes as PI, denoted
as \textsc{MIML-FCN+BB}, or image captions as PI, denoted as
\textsc{MIML-FCN+CP}. \item \textsc{G-MIML-FCN+}: Two stream
networks utilizing NN graphs. It also has two versions:
\textsc{G-MIML-FCN+BB} and \textsc{G-MIML-FCN+CP}. \squishend

\subsection{Settings and Parameters}
Following the discussions in Section~\ref{example}, we consider
each image from the datasets as a bag. For each image, we extract
maximum $500$ proposals using Regional Proposal Network
(RPN)~\cite{Ren2015}, each of which is considered as one instance
in the bag. This results in millions of training instances even
for the relatively small \textsc{VOC 2007} dataset.

For feature extraction, we utilize the network architecture of
Faster R-CNN~\cite{Ren2015}. Basically, our feature extraction
network is the VGG-16 network~\cite{Simonyan2014} with ROI pooling
layer, with the removal of all the classification / detection
related layers. For fair comparison, all methods we compare are
using these same features, although some methods likes our
\textsc{MIML-FCN+} and WSDDN~\cite{Bilen2016} can be integrated
with the feature extraction network and trained end-to-end.
\begin{figure}
	\centering
	{\includegraphics[width=0.45\textwidth]{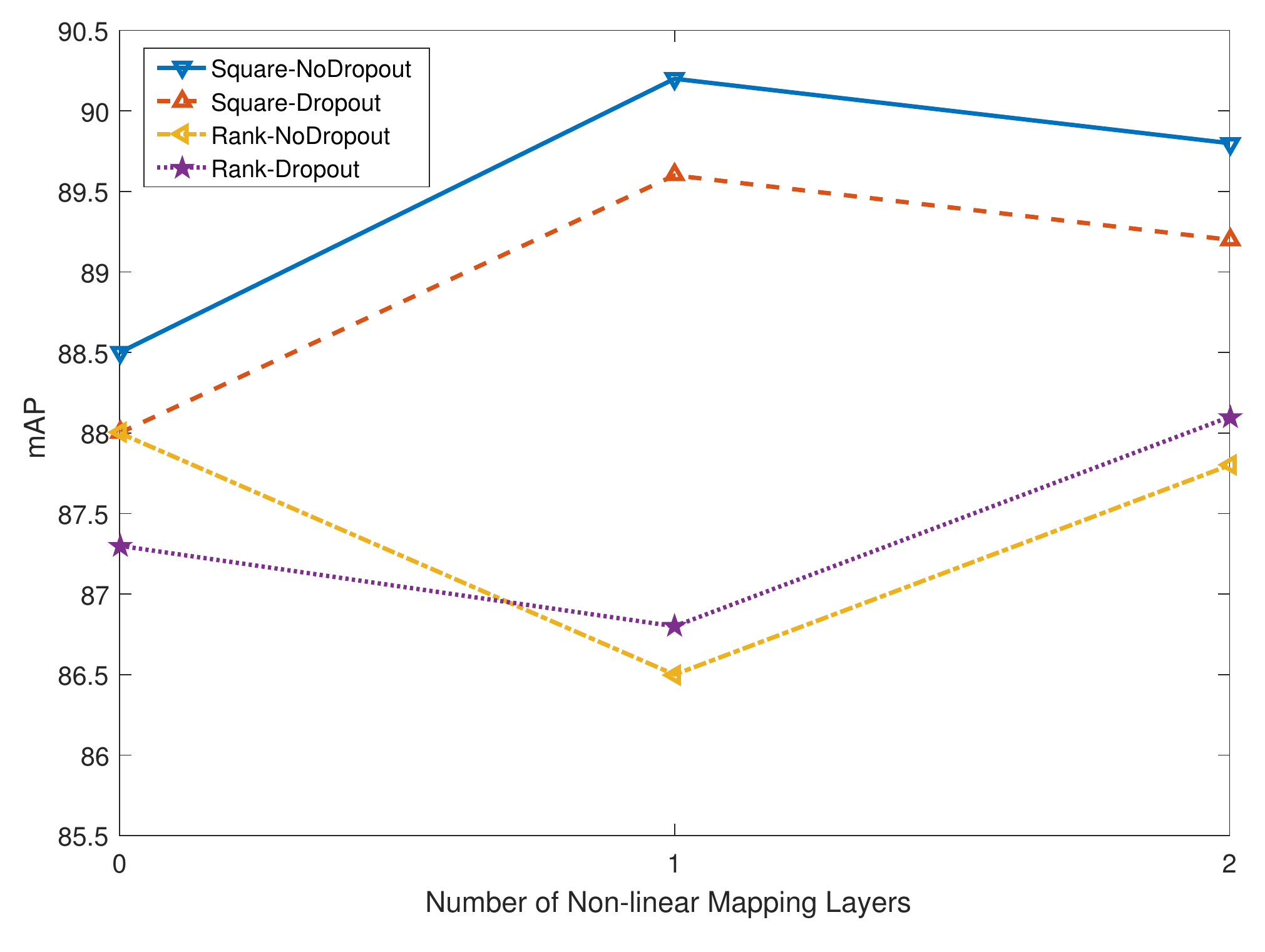}}
	\caption{Results of our MIML-FCN method under different network configurations on \textsc{VOC 2007} dataset. The $x$-axis represents the number of convolutional-ReLU layer-pairs as non-linear feature mapping. `Square-NoDropout': square loss without dropout; `Square-Dropout': square loss with dropout; `Rank-NoDropout': label ranking loss without dropout; `Rank-Dropout': label ranking loss with dropout.}
	\label{val}
\end{figure}

Our basic MIML-FCN consists of one convolutional layer, one ReLU
layer, one classification layer, one pooling layer and one loss
layer, as shown in Fig.~\ref{pi}. The convolutional layer contains
$2048$ filters in total. We tested a few possible numbers of filters, such as $\{4096,2048,1024\}$ and found out that $2048$ achieves slightly better accuracies. We also study the effects of different number
of convolution and ReLu layer-pairs, effects of dropout, as well
as the differences between square loss and label ranking loss. The
results are presented in Fig.\ref{val}. From these results, we
decide to choose one convolutional-ReLU layer-pair with square
loss.

Our main hyperprameter is the tradeoff parameter $\lambda$, which
is tune by cross-validation in a small subset of the training
data. The other important hyperparameter is the nearest neighour
number $k$ in \textsc{G-MIML-FCN+}, which we set to $5$ in all our
experiments. For other methods, we follow the parameter tuning
specified in their papers if available.

\subsection{Classification Results}
\begin{table}
	\small
	\centering \caption{Comparisons of classification results (in \%) of state-of-the-art approaches on \textsc{VOC 207}, \textsc{VOC 2012} and \textsc{MS COCO} datasets. The upper part shows results from other MIML learning methods, the middle part shows state-of-the-art recognition results and the lower part shows results from the proposed \textsc{MIML-FCN+} and its variations.}
	\begin{tabular}{ c | c | c | c}
		\hline
		&\textsc{VOC 2007} &\textsc{VOC 2012} &\textsc{MS COCO} \\ \hline
		\textsc{RankLossSIM\cite{Briggs2012}} &87.5 &87.8 &- \\
		\textsc{miFV~\cite{Wu2014}} &88.9 &88.4 &62.5\\
		\textsc{MIMLFast~\cite{Huang2014}} &87.4 &87.5 &61.5\\
		\hline
		\hline
		\textsc{WSSDN~\cite{Bilen2016}} &89.7 &89.2 &63.1 \\
		\textsc{VeryDeep~\cite{Simonyan2014}} &89.7 &89.3 &62.6\\
		\textsc{MVMI~\cite{Yang2016a}} &92.0 &90.7 &63.7\\
		\hline
		\hline
		\textsc{MIML-FCN} &90.2 &89.8 &63.5 \\
		\textsc{MIML-FCN+BB} &92.4 &91.9 &65.6\\
		\textsc{MIML-FCN+CP} &- &- &64.6\\
		\textsc{G-MIML-FCN+BB} &\textbf{93.1} &\textbf{92.5} &\textbf{66.2}\\
		\textsc{G-MIML-FCN+CP} &- &- &65.4\\
	\end{tabular}
	\label{res}
\end{table}
Table~\ref{res} reports our experimental results compared with
state-of-the-art methods on the three benchmark datasets.

Comparing our basic network \textsc{MIML-FCN} with
state-of-the-art MIML methods (upper part of the table), we can
see that our MIML-FCN achieves significantly better accuracies.
Specifically, \textsc{MIML-FCN} achieves around $1.2\%$
performance gain over miFV, which uses Fisher vector as a holistic
representation for bags. This suggests that using neural networks
for MIML problem can better encode holistic representation. One
interesting observation is that, if we remove the first
convolutional and ReLU layers of our \textsc{MIML-FCN}, it becomes
worse than miFV. This phenomenon confirms the effectiveness of
non-linear mapping component in our system. For \textsc{MIMLFast},
the main difference is that we employ square loss instead of label
ranking loss and we have a non-linear ReLU function. Our MIML-FCN
obtains more than $2\%$ accuracy gain over \textsc{MIMLFast},
which once again confirms the effectiveness of non-linear mapping
over linear mapping.

For comparisons with other state-of-the-art recognition methods
(middle part of the table), it can be seen that our basic MIML-FCN
achieves similar results as WSDDN, as the principles behind both
methods are similar. In contrast, instead of treating the task as
MIML problem, \textsc{VeryDeep}~\cite{Simonyan2014} just treats it
as multiple single label problems, where it uses multiple images
at different scales as network input, concatenates all the
features from different scales as the final representations and
then learns multiple binary classifiers from the representations.
Both our basic network \textsc{MIML-FCN} and \textsc{WSSDN}
achieve better performance than VeryDeep.

More importantly, Table~\ref{res} demonstrates the effectiveness
of using privileged information. Note that since captions are only
available in \textsc{MS COCO} dataset, MIML-FCN+CP is only applied
on COCO. From the table, we can see that \textsc{MIML-FCN+BB}
achieves around $2\%$ performance gain over MIML-FCN on all three
datasets, confirming the effectiveness our privileged bag idea.
Although \textsc{MIML-FCN+CP} is not as effective as
\textsc{MIML-FCN+BB}, it still outperforms \textsc{MIML-FCN}.
Comparing \textsc{MIML-FCN+BB} with the state-of-the-art
multi-view multi-instance (MVMI) framework~\cite{Yang2016a}, both
methods make use of bounding boxes, where our framework utilizes
BB as PI while their framework implicitly uses BB as label view in
the multi-view setup. Note that the results shown
for~\cite{Yang2016a} in Table~\ref{res} is a fusion of their
system and VeryDeep, but our \textsc{MIML-FCN+BB} still achieves
better performance.

In addition, comparing the results between MIML-FCN+BB and
G-MIML-FCN+BB and between MIML-FCN+CP and G-MIML-FCN+CP, we can
see that by further exploiting inter-instance correlations, our
framework can perform even better.

\section{Conclusion}
In this paper, we have proposed a two-stream fully convolutional
network, named MIML-FCN+, for multi-instance multi-label learning
with privileged bags. Compared with existing works on PI, we
explored privileged bags instead of privileged instances. We also
proposed a novel PI loss, which is similar to the high level idea
of SVM+, but is SGD-compatible and can be integrated into deep
learning networks. We have also explored the benefits of making
use of structured correlations among instances by simple
modifications to the network architecture. We demonstrated the
effectiveness of our system by a practical example of multi-object
recognition. We achieved significantly better performance in all
the three benchmark datasets containing millions of instances. For future directions, we intend to explore more possible applications
as well as other kinds of privileged information. We could also study the theoretical differences between the proposed PI loss and SVM+ loss.

{\small
	\bibliographystyle{ieee}
	\bibliography{egbib}
}
\end{document}